\documentclass{article}

\usepackage{PRIMEarxiv}

\usepackage[utf8]{inputenc} 
\usepackage[T1]{fontenc}    
\usepackage{hyperref}       
\usepackage{url}            
\usepackage{booktabs}       
\usepackage{amsfonts}       
\usepackage{nicefrac}       
\usepackage{microtype}      
\usepackage{lipsum}
\usepackage{fancyhdr}       
\usepackage{graphicx}       
\graphicspath{{media/}}     

\usepackage{amsmath,amsfonts}
\usepackage{algorithmic}
\usepackage{array}
\usepackage[caption=false,font=normalsize,labelfont=sf,textfont=sf]{subfig}
\usepackage{textcomp}
\usepackage{stfloats}
\usepackage[ruled,vlined]{algorithm2e} 
\usepackage{verbatim}

\title{Density-based Isometric Mapping
\thanks{
This is the authors' version.\\
B. Yousefi, M. Khansari, R. Trask, P. Tallon, C. Carino, L. Ma are with the University of Maryland, College Park, MD 20742. \\
A. Afrasiyabi is with Yale University, New Haven, CT 06520. \\
V. Kundra and L. Ren are with the University of Maryland School of Medicine, Baltimore, MD 21201.\\
K. Farahani is with the Center for Biomedical Informatics \& Information Technology, National Cancer Institute (NCI), National Institute of Health (NIH), Bethesda, MD 20814. \\
M. Hershman is with the University of Pennsylvania, and Boise Radiology Group, Philadelphia PA 19104.\\
The corresponding author is B. Yousefi, BSE 4117, 9636 Gudelsky Drive, Rockville, MD 20850, Tel: 240-665-6529, e-mail: byousefi@umd.edu\\}\\
}

\author{
  Bardia Yousefi, Mélina Khansari, Ryan Trask, Patrick Tallon, Carina Carino, Arman Afrasiyabi,\\
  \textbf{Vikas Kundra, Lan Ma, Lei Ren, Keyvan Farahani, Michelle Hershman }\\
}

\begin{document}
\maketitle

\begin{abstract}
The isometric mapping method employs the shortest path algorithm to estimate the Euclidean distance between points on High dimensional (HD) manifolds. This may not be sufficient for weakly uniformed HD data as it could lead to overestimating distances between far neighboring points, resulting in inconsistencies between the intrinsic (local) and extrinsic (global) distances during the projection. To address this issue, we modify the shortest path algorithm by adding a novel constraint inspired by the Parzen-Rosenblatt (PR) window, which helps to maintain the uniformity of the constructed shortest-path graph in Isomap.
Multiple imaging datasets overall of 72,236 cases, 70,000 MINST data, 1596 from multiple Chest-XRay pneumonia datasets, and three NSCLC CT/PET datasets with a total of 640 lung cancer patients, were used to benchmark and validate PR-Isomap. 431 imaging biomarkers were extracted from each modality. Our results indicate that PR-Isomap projects HD attributes into a lower-dimensional (LD) space while preserving information, which is visualized by the MNIST dataset indicating the maintaining local and global distances.  
PR-Isomap achieved the highest comparative accuracies of 80.9\% (±5.8) for pneumonia and 78.5\% (±4.4), 88.4\% (±1.4), and 61.4\% (±11.4) for three NSCLC datasets, with a confidence interval of 95\% for outcome prediction. Similarly, the multivariate Cox model showed higher overall survival, measured with c-statistics and log-likelihood test, of PR-Isomap compared to other dimensionality reduction methods. Kaplan Meier survival curve also signifies the notable ability of PR-Isomap to distinguish between high-risk and low-risk patients using multimodal imaging biomarkers preserving HD imaging characteristics for precision medicine.

\end{abstract}


\section{Introduction}
{H}{igh dimensional} data projection in lower space is always interesting since the data becomes sparse (in an exponential manner) by the \emph{curse of dimension}. Hence, several different dimension reduction (DR) ~\cite{r1,r2,r3,phate} have been proposed under the machine learning paradigm to visualize or build a model with better generalization performance. Within this context, a viable approach to streamline data analysis involves the assumption that the dataset in question inherently resides within a lower-dimensional (LD) space. Data characterized by lower dimensions can be conveniently visualized within this reduced-dimensional framework, rendering them more amenable to analysis and interpretation. DR techniques play a pivotal role in this process, as they facilitate the transformation of data between the HD space and the LD embedding, serving as a bridge between these two domains.
Principal component analysis (PCA) ~\cite{r1} - also called Karhunen–Loève theorem – KLT is a representation of a stochastic process as an infinite linear combination of orthogonal functions, analogous to a Fourier series representation of a function on a bounded interval. This is modified by singular value decomposition (SVD) and factor analysis. In particular, PCA is a DR technique that effectively preserves the pattern of data in a linear subspace. 
However, in many instances, data exhibits a nonlinear manifold that cannot be geometrically represented by PCA. Here, manifolds are geometrical structures that can be resided with a few degrees of freedom in the latent space. Manifold models can correctly symbolize the association of data points in physical systems or be controlled by a small number of continuous projected parameters. Any point on the manifold can be profoundly retained by LD vectorized representation. The overarching goal of the manifold learning methods is to acquire appropriate non-linear embedding methods in the LD space that preserve HD data configuration to simplify data analysis and prevent issues with HD data. 
Alas, most of the well-known DR methods such as multidimensional scaling (MDS) ~\cite{r2}, isometrically mapping (Isomap) ~\cite{r3}, locally linear embedding (LLE) ~\cite{r4}, Laplacian eigenmaps ~\cite{r5}, Hessian eigenmaps ~\cite{r6}, Sammon mapping ~\cite{r7}, and graph embedding ~\cite{r8} are inherently predicated outside of a stringent uniformity assumption concerning the underlying data distribution. Therefore, these methods are vulnerable to nonuniform data spread in the manifold. However, in real-world applications, we may encounter non-uniform data, particularly in medical data, which may affect the diagnostic/prognostic outcomes.

Unsupervised methods within the spectral approach are Sammon~\cite{r7} and Isomap~\cite{r3}, which are related to MDS ~\cite{r9} and attempt to nonlinearly preserve the similarities and distances, respectively. Isomap uses geodesic distance instead of Euclidean distance in the formulation of the kernel ~\cite{r3}. Isomap is classified as a non-linear method owing to its reliance on geodesic distance.
An important point to notice is that Isomap uses geodesic distance to stay on the manifold for the calculation of the distance between two points far on the curved manifold ~\cite{r3}. This discriminates Isomap from classical MDS, where Euclidean distance was calculated for the points, and creates problems once points are distant from each other. With geodesic distance in the Isomap, there will be many smaller distances selected and approximated with Euclidean distances with the k-nearest neighbor path defined on the manifold. Standard Isomap considers the k-nearest neighbor for finding the shortest path with specific k neighbors using Dijkstra’s algorithm. There is no guarantee that these neighbors are close enough to each other so an approximation of geodesic with Euclidean distances stays on the manifold. In other words, if we select large k, there will be points that are far enough on the manifold that Euclidean distance would not be calculated on the manifold and imitates the same issue that the MDS faces.
To relax the discussed data uniformity assumption and enforce the geodesic points to stay on the manifold we propose a constraint Parzen–Rosenblatt (PR) window ~\cite{r13,r14} -based method which constraint to modify local neighbors’ graph to highlight the uniformity of the data distribution on the HD manifold for isometric mapping, PR-Isomap, which preserves both distances and uniformity criteria while nonlinearly reducing the dimensionality.
In terms of distant points on curved manifolds, PR-Isomap limits the number of neighbor points selected on the manifold’s surface. This overcomes the issue of far points on the manifold that standard Isomap does not have, which is considered the main novelty of our approach. With PR-Isomap there is more guarantee of approximating geodesic distance between two far points into many smaller Euclidean distances on the manifold. This modifies and simultaneously maintains the original assumption of isometric mapping ~\cite{r3}. 
	
\begin{figure}[t]
\begin{center}
   \includegraphics[width=0.65\linewidth]{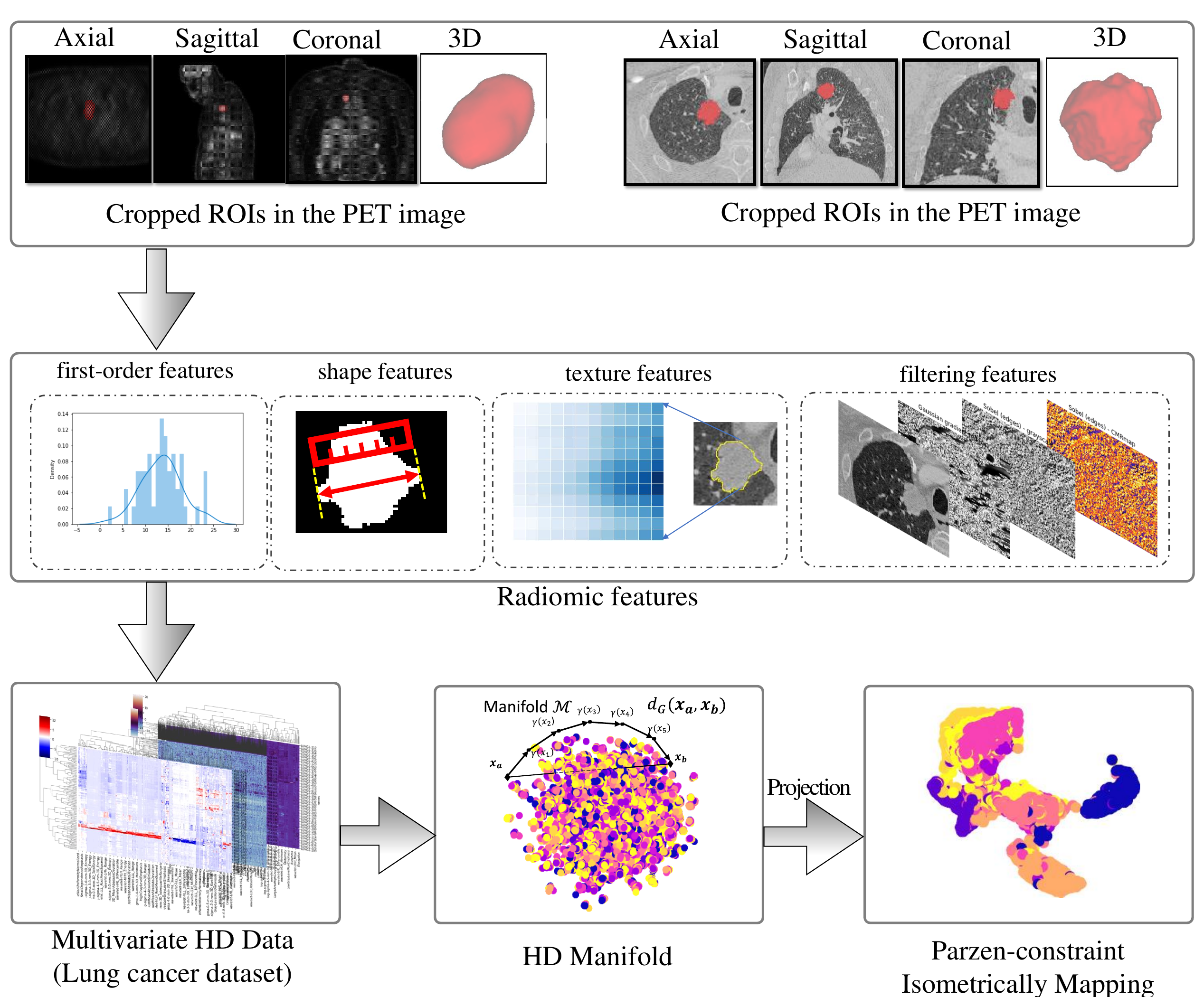}
\end{center}
   \caption{Our PR-Isomap method deflates a high-dimensional manifold onto a low-dimensional representation while preserving their similarities. We generate the HD-manifold using NSCLC Radiogenomics dataset ~\cite{r29,r30,r31}. We apply the standard t-SNE ~\cite{r11}, Isomap ~\cite{r3}, PCA ~\cite{r1}, and our proposed PR-Isomap method to reduce the dimensionality of the data. PR-Isomap preserves similarity in low-dimensional embeddings and showed more promising results in predicting patients’ survival.}
\label{fig:one}
\label{fig:onecol}
\end{figure}

As Fig. \ref{fig:one} illustrates, our dimension reduction method can extend to medical data analysis investigation, where we generate the HD manifold using multimodal imaging from the lung cancer NSCLC Radiogenomics dataset ~\cite{r29,r30,r31}. We previously applied data reduction for NSCLC prognosis for liquid biopsy data integration (~\cite{y100,y101}), NSCLC intraobserver variability~\cite{y102}, and breast cancer diagnosis while reducing data hierarchically between temporally and for features~\cite{y103,y104}. Particularly, we apply the standard t-SNE, PCA, PHATE (Potential of Heat-diffusion for Affinity-based Trajectory Embedding) ~\cite{phate}, standard Isomap, and our proposed PR-Isomap method to reduce the dimensionality of the data. PR-Isomap preserves distance in LD embedding and showed more promising results in predicting patients’ survival while it is tested for 1,596 cases of Pneumonia and 70,000 MNIST data. 
In our PR-Isomap, for preserving local distances for the formation of data, we adopt a similar concept as in the original Isomap to keep the geodesic distance of pairwise points on the HD manifold. This unfolds the manifolds and makes it easier to project data points to the LD space. We adopt PR window formation and use it as a constraint for local neighbors to emphasize the continuity of data points on the manifold while calculating geodesic distance. Overall, this paper presents the following contributions:
\begin{list}{-}{}
\item{We propose a novel DR method, PR-Isomap, which can project HD data points into LD space while adeptly conserving both local and global distances, ensuring the preservation of the overall data structure.}

\item{We modify Isomap to use the PR window constraint in the local neighbors’ graph to emphasize the data’s uniformity distribution while computing LD similarities which is critical to maintaining uniformity and maintaining approximation of geodesic distance to many smaller Euclidean distances on the manifold.}

\item{We conduct a comparative analysis of known state-of-the-art DR methods and reduce HD data, and multimodal radiogenomic dataset and use the latent space representation of these data to predict the survival of patients under cancer treatment, Pneumonia diagnosis, and visualizing the MNIST dataset. These applications were conducted to better understand the strengths and weaknesses of the proposed approach. Throughout all the circumstances, PR-Isomap delivers better results than the baselines.}
\end{list}
In the next section, we discuss related works in DR and Isomap. In the section after that, our DR method to tackle HD imaging biomarkers and constraint isometric mapping is presented (Section 3). The experimental and computational results, as well as the discussion, are then presented in Sections 4 and 5, respectively. Finally, the conclusions are presented in Section 6.

\section{Related Work}
\noindent Unsupervised DR techniques can generally be divided into two categories:
\begin{list}{}{}
\item{ {\bf{LD Feature extraction:}} The goal here is to transform the original HD data into the LD space by extracting new features that are a linear or nonlinear combination of the original features, e.g., principal component analysis (PCA)  ~\cite{r1}. PCA effectively preserves the pattern of data in a linear subspace. However, in many instances, data exhibits a nonlinear manifold that cannot be geometrically represented by PCA due to the linearity of the approach.}\\

\item{ {\bf{Manifold learning:}} These methodologies endeavor to unveil an LD manifold that is embedded in HD data and captures the fundamental structure of the data. Examples of manifold learning techniques include MDS  ~\cite{r2}, Isomap ~\cite{r3}, LLE ~\cite{r4}, Hessian eigenmaps ~\cite{r6}, Sammon mapping  ~\cite{r7}, and graph embedding ~\cite{r8}. Manifold learning can be broadly divided into two categories based on the type of manifold structure that is assumed to be present in the data.}\\

\item{ {\bf{Global manifold learning:}} The assumption here is that the data can be represented by a solitary, comprehensive LD manifold that encapsulates the fundamental structure of the data. Illustrations of global manifold learning methods consist of Isomap and Laplacian Eigenmaps.}\\

\item{ {\bf{Local manifold learning:}} Such techniques operate under the assumption that the data can be modeled by a set of local LD manifolds, each capturing the structure of the HD data, i.e., LLE ~\cite{r4} and Hessian Eigenmaps ~\cite{r6}. LLE aims to reconstruct each data point linearly using its neighbors in the original space, and it employs the same weights of linear reconstruction for embedding in the LD space. Meanwhile, Kernel LLE conducts the stages of finding neighbors and linear reconstruction in the feature space.}
\end{list}

\noindent Notwithstanding their common objective of uncovering the intrinsic structure of data in an LD space while preserving the critical characteristics of HD data, manifold learning techniques are differentiated by their underlying manifold assumptions, criteria, and methods of constructing the LD representations.

\textbf{Similarity criteria for DR methods:} t-distribution stochastically embedding (t-SNE)~\cite{r11} use Kullback-Leibler (KL) divergence between the two joint probability distribution and often used for visualization without emphasizing on uniformity of data distribution. Uniform manifold approximation and projection (UMAP) is the only method that focuses on the uniformity of data distributions on manifolds applying Riemannian geometry and category-theoretic approaches with fuzzy simplicial sets ~\cite{r20}. UMAP exhibits a lack of strong interpretability despite significant performance.

\textbf{Distance metric-based DR methods:} Sammon mapping, PHATE (Potential of Heat-diffusion for Affinity-based Trajectory Embedding)~\cite{phate}, and Isomap are related to MDS  ~\cite{r2}, attempting to nonlinearly preserve the pairwise similarities and distances, respectively. LLE ~\cite{r4} and Kernel LLE ~\cite{r18} also use distance measures to find neighbors and linear reconstruction in the feature space. Isomap uses geodesic distance rather than Euclidean distance in the kernel formulation ~\cite{r3}.

\textbf{Isometric mapping and its variations:} Isomap endeavors to conserve the intrinsic structure of the data. However, its intricate computation presents a formidable obstacle when it comes to processing large datasets. Several Isomap variants exist to address its computational complexity for large datasets. One such variant is C-Isomap ~\cite{r50}, which assigns different weights to neighborhoods based on the density of data points, providing a more accurate representation of the underlying data structure. Similarly, C-Isomap has a computational complexity of $O(N^3)$, making it challenging to use for large datasets. To address the issue, another variant of Isomap selects a small set of $M$ landmark points ($M << N$) to compute the geodesic distances between each landmark and the rest of the data points, reducing the computational complexity to $O(M^3)$, making it more manageable for large datasets.

Angle constraints are applied to the shortest path to ensure the smoothness of the paths, and to accurately estimate geodesic distances, even in situations where the surface intersects itself ~\cite{r23}, which does not focus on the uniformity data points. Spatiotemporal Isomap (ST-Isomap) ~\cite{r17} analytically modifies the local neighbors’ graph weights to highlight the similarities to temporally related observations.
Our PR-Isomap method, as proposed, can be effectively employed in HD manifold embeddings. It enhances the emphasis on uniformity, preserving both pairwise distances and the overall uniformity of the data.

\begin{figure}[t]
\begin{center}
   \includegraphics[width=0.75\linewidth]{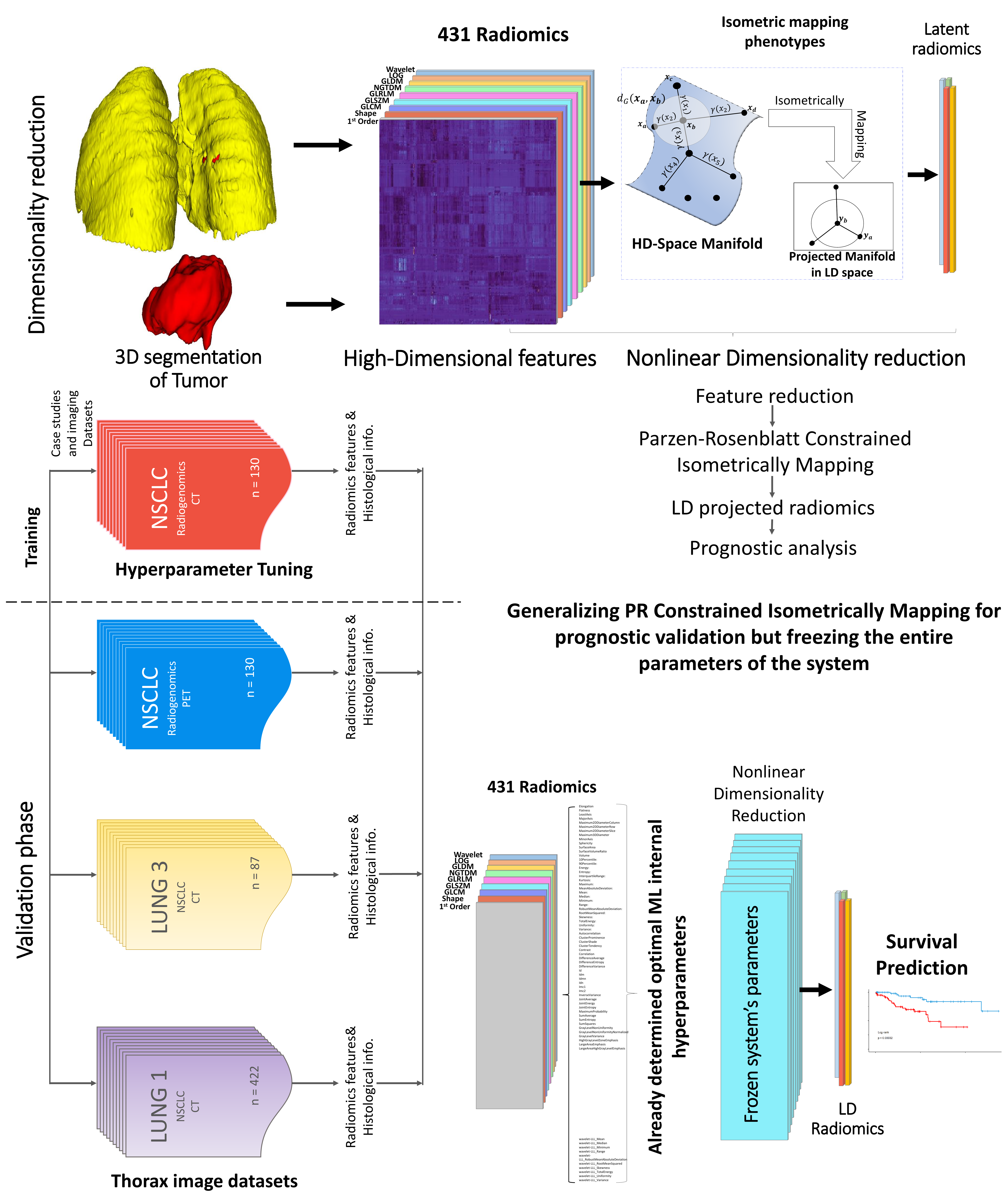}
\end{center}
   \caption{Workflow of the proposed approach. A constrained isometrically mapping method is used to reduce feature dimensionality. Then, a subset of radiomic signatures is used to predict the survival of the patients using the Cox proportional hazard multivariate model. Afterwards, the parameters of the system were frozen, and the same analyses in an independent NSCLC dataset.}
\label{fig2}
\end{figure}

\section{Materials and Methods}
\noindent In the following sections, we introduce the isometric mapping method and discuss the innovative techniques used to extract maximum LD information for predicting survival and classification tasks (Fig. \ref{fig:onecol}).
\vspace{-.1in}
\subsection{Isometrically Mapping DR}
\noindent Isomap is closely linked to the standard MDS, which seeks to identify underlying dimensions that may help to describe differences among HD features.
To deal with nonlinear complicated data, there are solutions to be followed. Here, one way is to use kernel methods to project data to the HD space where we can linearly group data. Another approach is to use nonlinear methods to unfold the nonlinearity in the data. Isomap uses geodesic distance to unfold the HD manifolds to a simpler one and makes them easier to project to LD space.
Our data points can be shown as a combination of multiple vectorized data points, $X = \left[ \mathbf{x}_1, \mathbf{x}_2, \ldots, \mathbf{x}_n \right]\in \mathbb{R}^{d \times n}$, where $\big\{ {\mathbf{x}_i\in\mathbb{R}^d} \big\}_{i=1}^n$. We assume that there is a finite data point drawn from our X. We can build a geometric graph $G = (V;E)$ that has our data points as vertices and connects vertices that are close to each other. We construct a k-nearest neighbor graph for the $\textbf{x}_i$ connected to $\textbf{x}_j$ correspondingly and for $\textbf{x}_j$ to $\textbf{x}_i$, show them by $\left(\mathbf{x}_i\right)\ $ and $ \gamma\left(\mathbf{x}_j\right)$, $ \gamma\left(\mathbf{x}_i\right),\ \gamma\left(\mathbf{x}_j\right) \in \mathcal{M}$. $ \gamma(.)$ represents a vector point on the manifold. The length of the path is defined by summing the edge weights along the path on the manifold. The shortest path ($\mathcal{SP}$) distance $D_{\mathcal{SP}}(\gamma\left(\mathbf{x}_i\right),\ \gamma\left(\mathbf{x}_j\right))$ between two vertices $ \gamma\left(\mathbf{x}_i\right),\ \gamma\left(\mathbf{x}_j\right)\in V$ and defines based on the length of the $\mathcal{SP}$ connecting them. 
$g\left(\cdot \right)$ is a positive continuous function for $X$ in the given path $\gamma(\cdot )$ connecting $ \mathbf{x}_i\ $ and $\mathbf{x}_j$, which is parameterized by $g$-length of the path as
\begin{equation}
    D_{g,\gamma} (t) = \ \int_{\gamma}{g\left(\gamma \left(t\right)\right)\ \left|\ \gamma^\prime\left(t\right)\right| \ \ dt},
\end{equation}
where $\gamma^\prime(\cdot )$ defines a small element on the manifold, also known as the line integral along the path $\gamma(\cdot)$ with respect to $g(\cdot)$. $ D_{g,\gamma}\ $ denotes as distance on the manifold through the shortest path. The g-geodesic path connecting $ \mathbf{x}_i\ $ and $\mathbf{x}_j$ is the path with minimized g-length ~\cite{r21}. The shortest path is used for the Isomap algorithm to find the path on the HD manifold for connecting two points.
We assume that the LD embedding of the training data with respect to the vectorized data points of the observation can be shown as $Y = \left[\mathbf{y}_1,{\ \mathbf{y}}_2,\ldots,\mathbf{y}_n\right]\in\mathbb{R}^{p\times n}$, where $\big\{ {\mathbf{y}_i\in\mathbb{R}^p} \big\}_{i=1}^n$, and $p\le d$, desirably and usually $p\ll d$. The MDS method aims to project similar points of the data close to each other and dissimilar points as far as possible. This is followed by the equation below:

\begin{equation}
\begin{aligned}
\underset{Y}{\text{min}} \| D_{g,\gamma}(X)-D_{g,\gamma}(Y)\|_F^2,   \\
\end{aligned}               
\end{equation}
where $D_{g,\gamma} (\cdot)$ is the shortest path distance from equation (1). $D_{g,\gamma}\left(X\right)\ $ and $D_{g,\gamma}(Y)$ are any pairwise distances of the data in HD space and in the subspace and can be written by similarity $X^TX$ and $Y^TY$ in the projected space, respectively. In MDS, the solution for this problem is $Y=\Delta^\frac{1}{2}{\ V}^T$, where $V^T$ and $\Delta\ $ are the eigenvector and eigenvalue matrices of the input data, $X$. This is equivalent to the PCA method. If we use a generalized kernel with respect to the kernel distance matrix $D$ as:
\begin{equation}
\begin{aligned}
    K= \ -\frac{1}{2}HDH
\end{aligned} 
\end{equation}
with the centering matrix of $H$, $H=I-\frac{1}{n}1{1}^T$, where $I$ is centering matrix and $1$ are $[1,\dots,1]^T$ identity matrix. In generalized MDS embedding we use eigen decomposition of $K$. The MDS applies Euclidean distances between data points as a similarity measurement. In HD space with curves, i.e.,  non-Romanian manifolds, the Euclidean distance may not serve as an accurate measure of the distance between the points on the manifold. However, it's worth noting that the intrinsic and extrinsic distances between two points on the curved manifold may differ. In other words, Euclidean distance may not be an accurate representation of the true distance between points in such spaces. Isomap solved this problem ~\cite{r3} by using the geodesic distance between the points on the HD manifold. The Isomap algorithm comprises two main steps:

1. Creating a k-nearest-neighbor graph in HD space of the points on the manifold, and then calculating the geodesic distance between each pair of points applying shortest-path distance.

2. Employ MDS and make the distance matrix to locate the points in the embedded LD space. Isomap follows the Cayton assumptions ~\cite{r22,r23} to retrieve the manifold parametrization:
\begin{list}{}{}
\item{i. An isometric embedding of the manifold in $ \mathbb{R}^p $ exists.}
 
\item{ii. The manifold in $\mathbb{R}^d$ is convex as well as its parameter space. In other words, there is no hole in the manifold. Also, the geodesic distance between any two points in the parameter space remains entirely in the parameter space.}
 
\item{iii. Everywhere on the manifold, points are sufficiently sampled.}

\item{iv. Maintain the compactness of the manifold topologically.}
\end{list}
Increasing the distance between the paired points mitigates the similarity of these two distances and the accuracy of such an approximation, which ultimately leads to the original MDS problem. 

\vspace{-.1in}
\subsection{Constrained Isomap }
\noindent The Isomap algorithm creates a k-nearest neighbor graph and uses the shortest path algorithm to estimate the geodesic distances between points. After finding geodesic distances between points, it searches for geodesic distances equivalent to Euclidean distances. Due to isometrically embedded manifold conditions, some points satisfy these conditions along with translation and rotation, e.g., MDS is a classic example of finding these points ~\cite{r24}. 
This study focuses on the Isomap algorithm for nonlinear manifold learning, which aims to maintain the pairwise geodesic distances between data points ~\cite{r3,r25,r26}. To calculate the geodesic distance and the shortest path between two data points, we make the assumption that the data points belong to a $d$-dimensional manifold that is embedded in an $ \mathbb{R}^d $ ambient space. The crucial assumption for the Isomap method is the presence of an isometric chart, in which pair points such as $\mathbf{x}_i,\mathbf{x}_j$ lie on the HD manifold. $D_{g,\gamma}\left(\mathbf{x}_i,\mathbf{x}_j\right)\ $ denotes the geodesic distance of the pairwise points  $\gamma:\mathcal{M}\rightarrow\ \mathbb{R}^d$ such that:
\begin{equation}
    D_{Euc}\left(\mathbf{x}_i,\mathbf{x}_j\right)\ \propto \| \gamma\left(\mathbf{x}_i\right)-\gamma\left(\mathbf{x}_j\right) \|_2
\end{equation}
where $ \mathbf{x}_i\ $ and $\mathbf{x}_j$ are near each other on the manifold. Geodesic distance of $\gamma\left(\mathbf{x}_i\right)\ $ and $ \gamma\left(\mathbf{x}_j\right), (\gamma\left(\mathbf{x}_i\right),\ \gamma\left(\mathbf{x}_j\right)) \in  \mathcal{M}$ defines along the manifold surface, $ \mathcal{M}$, and is denoted by $D_{Euc}\left(\mathbf{x}_i,\mathbf{x}_j\right)$ and can be approximate by $ D_{g,\gamma}\left(\mathbf{x}_i,\mathbf{x}_j\right)\ $.  

\begin{figure}[t]
\begin{center}
   \includegraphics[width=0.55\linewidth]{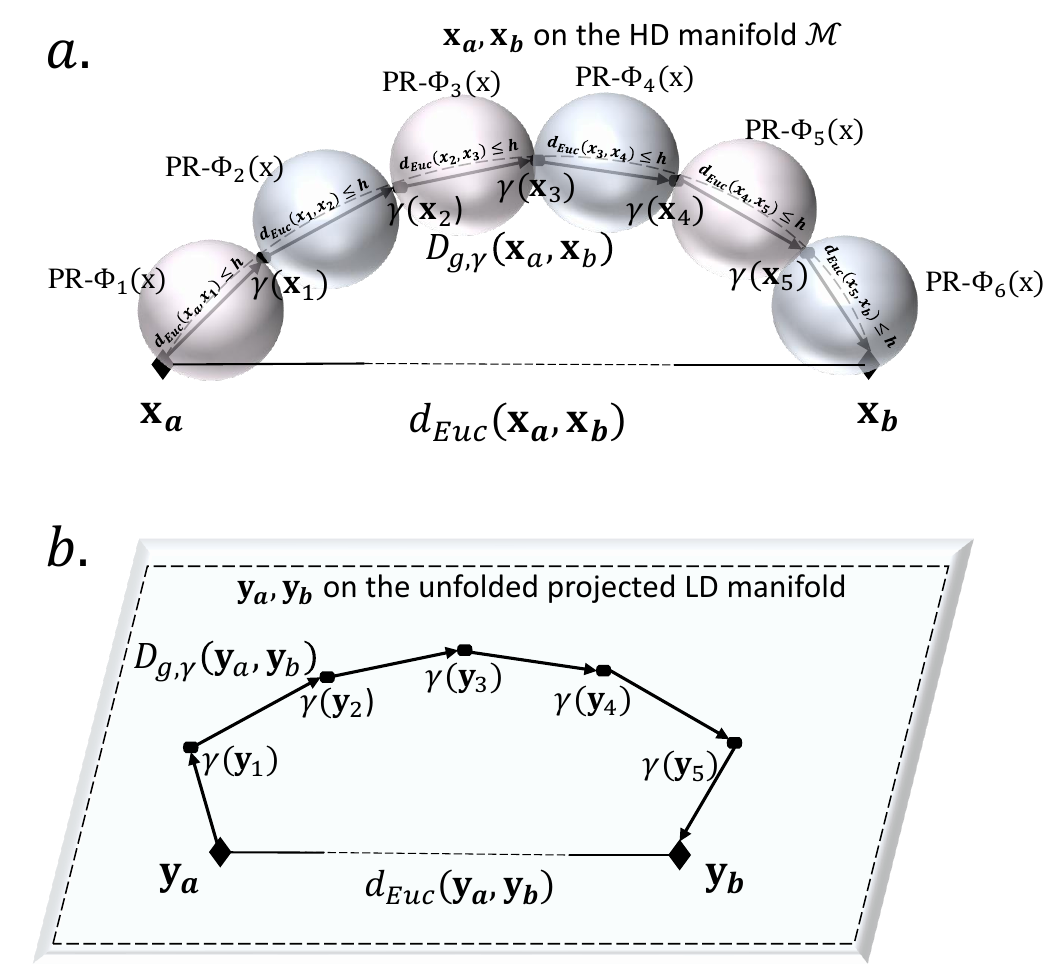}
\end{center}
   \caption{Geometric interpretation of the shortest path and PR-Isomap under $\mathcal{N}_{k,h}\left(\gamma\left(\mathbf{x}_k\right)\right)$, Parzen–Rosenblatt window constraint on $k$-nearest neighbor windows on HD manifold (a) and in the LD space after PR-Isomap projection (b). $d_{Euc}(\mathbf{x}_a,\mathbf{x}_b)$ and $D_{g,\gamma}\left(\mathbf{x}_a,\mathbf{x}_b\right)\ $are Euclidian and geodesic distances between $\mathbf{x}_a$ and $\mathbf{x}_b$.}
\label{fig:long}
\label{fig:3}
\end{figure}

By building a neighborhood graph where each point is connected exclusively to its k nearest neighbors, Isomap first calculates an estimate of the geodesic distances between every pair of data points located on the manifold $ \mathcal{M}$; the edge weights are equivalent to the corresponding pairwise distances. The Euclidean distance serves as an estimate for the geodesic distance for adjacent pairs of data points. This assumption allows finding an approximation of $D_{g,\gamma}\left(\mathbf{x}_i,\mathbf{x}_j\right)$ applying many small Euclidean distances, e.g.,
\begin{equation*}
    D_{g,\gamma}\left(\mathbf{x}_i,\mathbf{x}_j\right) \approx \| \gamma\left(\mathbf{x}_i\right)-\gamma\left(\mathbf{x}_j\right) \|_2  \ \ 	\forall  \ \ \gamma\left(\mathbf{x}_j\right)\in\mathcal{N}_k\left(\gamma\left(\mathbf{x}_j\right)\right),
\end{equation*}
where $ \mathcal{N}_k\left(\gamma\left(\mathbf{x}_j\right)\right)$ denotes the collection of the point $ \gamma\left(\mathbf{x}_j\right)\in\ \mathcal{M}$ for $k$ closest neighbors on the manifold surface, 
$ \mathcal{M}$. The geodesic distance is calculated for non-neighboring points as the shortest path length along the neighborhood graph, which can be discovered using Dijkstra's algorithm. The generated geodesic distance matrix is then subjected to MDS ~\cite{r2} to identify a group of LD points that most closely match such distances.
Forming a k-nearest neighbor graph for the data points might create inconsistency between geodesic and Euclidean distances in opposition to the aforementioned condition. Particularly, if the third Cayton assumption does not or is weakly satisfied, it indirectly implies that points in the HD manifold should be uniformly distributed. If otherwise, for large $k$ in the k-nearest neighbor graph there will be cases that create a discrepancy in an approximation of the geodesic and Euclidean distances. This will cause a difference between intrinsic and extrinsic distances on the manifold.
Here, we tackle this problem using the \textit{Parzen–Rosenblatt} window constraint on the k-nearest neighbor for a weak-uniformly distributed manifold which enforces the accuracy of distance approximation in the Isomap algorithm (Algorithm \ref{algo:PR-Isomap} and Fig. \ref{fig:3}). In this respect,
Parzen–Rosenblatt (PR) window is defined as
\begin{equation}
    p_h\left(\mathbf{x}\right)=\frac{1}{k}\sum_{i=1}^{k}{\frac{1}{h^2}\Phi\Big(\frac{\mathbf{x}_i-\mathbf{x}}{h}\Big)},
\end{equation}
where $k$ is the number of neighbors centered in the vector, $x$, $x$, $p_h(\mathbf{x})$ denotes the probability density of $\mathbf{x}$, while $h$ is the diameter of the window that helps to satisfy the approximation of geodesic and Euclidean distances, and $\Phi$ represents the window function, This could be a rectangular function or a uniform distribution, similar to a window of size $h$. This function is also called the density estimation function.
In addition, PR-Isomap alters the graph weights of a portion of each point's k-nearest neighbors with respect to the pairwise distances on the surface of the manifold, $\mathcal{M}$, that meet the requirements for the PR window. The set of a PR-limited points $ \gamma\left(\mathbf{x}_i\right), \gamma\left(\mathbf{x}_i\right)\in\ \mathcal{M}$ are considered for projection as neighboring elements. Suppose that the point $ \gamma\left(\mathbf{x}_j\right)$, where 
$ \gamma\left(\mathbf{x}_i\right)\in\ \mathcal{N}_{k,h}\left(\gamma\left(\mathbf{x}_j\right)\right)$ is the closest matching point that can be used for measuring the distance $D_{g,\gamma}\left(\mathbf{x}_i,\mathbf{x}_j\right)$, i.e.,
\begin{equation}
\begin{aligned}
& \underset{\gamma\left(\mathbf{x}_k\right)\in\ \mathcal{N}_{k,h}\left(\gamma\left(\mathbf{x}_k\right)\right)}{\text{min}}
& & D_{g,\gamma}\left(\mathbf{x}_i,\mathbf{x}_j\right) . \\
\end{aligned}
\end{equation}
PR-based isometric mapping uses a subset of k-nearest neighbor with a distance less than $h$ on the surface of the manifold, $\mathcal{M}$, which is represented by uniform neighbors $(\mathcal{UN}(\mathbf{x}_i))$, i.e.,
\begin{equation*}
    \mathcal{UN}\left(\mathbf{x}_i\right)= \{ \gamma\left(\mathbf{x}_i\right)\in\mathcal{M},  \ \ \gamma\left(\mathbf{x}_i\right)\in\ \mathcal{N}_{k,h}\left(\gamma\left(\mathbf{x}_i\right)\right), D_{g,\gamma}\left(\mathbf{x}_i,\mathbf{x}_j\right) \le  
    \begin{aligned}
& \underset{\gamma\left(\mathbf{x}_k\right)\in\ \mathcal{N}_{k,h}\left(\gamma\left(\mathbf{x}_k\right)\right)}{\text{min}}
& & D_{g,\gamma}\left(\mathbf{x}_i,\mathbf{x}_j\right) \\
\end{aligned}   \}
\end{equation*}\\
In other words, $ \mathcal{UN}\left(\mathbf{x}_i\right)$ is a subset of the nearest neighbor graph, which falls into the PR radius, containing points on the manifold $ \mathcal{M}$ and they are close to the targeted point out of the entire trivial match. $ \mathcal{UN}\ $ points are employed to detect data points on the local neighborhood of every point $ \gamma\left(\mathbf{x}_i\right)\ $ with a higher expectation to be analogous to $ \gamma\left(\mathbf{x}_i\right)$. 
Using the MDS generalized optimization:
\begin{equation}
    \begin{aligned}
     & \underset{Y}{\text{min}}
    & &  \| D_{g,\gamma}(X)-D_{g,\gamma}(Y)\|_F^2 .  \\
    & \text{subject to}
    & & \gamma\left(\mathbf{x}_k\right)\in\ \mathcal{N}_{k,h}\left(\gamma\left(\mathbf{x}_k\right)\right)
\end{aligned}
\end{equation}


\begin{algorithm}[t]
    \SetAlgoLined
    {\footnotesize
        \KwData{ \{$\mathbf{x}_1, \mathbf{x}_1, \dots, \mathbf{x}_n \} \in \mathbb{R}^D$.}
        \KwResult{$\{ \mathbf{y}_1, \mathbf{y}_1, \dots, \mathbf{y}_n \} \in \mathbb{R}^d$.}
        \textbf{Estimate: }{MDS on $D$- Geodesic distance pairwise point via shortest path \& neighborhood graph dissimilarity matrix $D$. $D \in \mathbb{R} ^{n \times n} (D_{ii} = 0, D_{ij} \geq 0)$}, $K=\ -\frac{1}{2}HDH$\\
        $K=\ -\frac{1}{2}HDH \:$ centering matrix of $H$, $H=I-\frac{1}{n}{11}^T$\\
        $Y=\Delta^\frac{1}{2}{\ V}^T$\\
        \textbf{Constrained Dijkstra’s Algorithm}\\
        Form a weighted undirected graph $\mathcal{G}_e = (\nu_e, E_e, w_e)$ for q-nearest neighbor, $q \leq k$. \\
        For $\forall$ vertex $v$, $v \in \mathcal{G} : D_{\gamma}[v] \leftarrow \infty,$  
        $parent[v] \leftarrow [], D_{\gamma}[s]] \leftarrow 0$, \\
        $Q := \forall \:$ nodes,$n$, $n \in \mathcal{G}$ \\
        
        \While{$Q \neq []$} {
            $u \leftarrow v \in Q \: \& \: \min D_{\gamma}[u]$\\
        remove $u$ from $Q$\\
            \For{{$\forall \: \mathcal{N}_q[u,v] \in Q$}}{
                \If{$D_{\gamma}[u,v] < h ,\:\: u,v \in \mathcal{N}_{k,h}$}{
                alt $\leftarrow D_{\gamma}[u] + \mathcal{G}_e edge(u,v)$\\
                    \If{alt $<$ $D_{\gamma}[v]$}{
                    $D_{\gamma}[v] \leftarrow$ alt \\
                    $parent[v] \leftarrow u$ \\
                    }
                }
            }                   
        }
    \caption{PR-Isometric Mapping.}
    \label{algo:PR-Isomap}}
\end{algorithm} 

  \emph{\textbf{Definition 1:}} Consider a geometric graph based on a fixed set of points $ \mathbf{x}_1,\ldots,  \mathbf{x}_2\in \mathbb{R}^d$. Let $h$ be a real number defined for $ \mathcal{N}_{k,h}\left(\gamma\left(\mathbf{x}_k\right)\right)$ such that $D_{g,\gamma}\left(\mathbf{x}_i,\mathbf{x}_j\right)\ \le h $ implies that $ \mathbf{x}_i$ is connected to $ \mathbf{x}_j$ on the graph.

Using Definition 1, if we rewrite equation (7) instead of uniform neighboring points on the HD and the LD space, but with a difference in pairwise distances in the HD space as a constraint:
\begin{equation}
    \begin{aligned}
     & \underset{Y}{\text{min}}
    & &  \| D_{g,\gamma}(X)-D_{g,\gamma}(Y)\|_F^2   \\
    & \text{subject to}
    & & D_{g,\gamma}\left(\mathbf{x}_i,\mathbf{x}_j\right)\le h
\end{aligned}
\end{equation}
where $D_{g,\gamma}(X)\ $ redefines as any pairwise distances on the surface of the HD manifold, $ \mathcal{M}$, while the $ \mathcal{SP}$ distance using an unweighted PR-based k-nearest neighbor graph, $p_h$. Fig. \ref{fig:3} presents the geometrical interpretation of the shortest path and PR-Isomap under $ \mathcal{N}_{k,h}\left(\gamma\left(\mathbf{x}_k\right)\right)$, where PR-window constraints the k-nearest neighbor path.\\
To evaluate the strength of LD features, we use multiple classification scenarios i.e., Random forest, and the Cox proportional hazard model to predict patients’ Pneumonia and survival, respectively.

\vspace{-.1in}
\subsection{Outcome prediction and statistical analysis}
\noindent In the proposed model, statistical analysis of image intensity, texture, contextual information corresponding to intratumor, and ROIs are obtained and listed as HD imaging attributes. These HD radiomics spanned onto LD space while their maximum information is preserved. To evaluate the performance of LD radiomics in predicting overall survival (OS), we performed binary classifications to predict treatment failure. 
Similar to the PR-Isomap method, we applied other commonly used state-of-the-art unsupervised DR methods to the same data as a comparison analysis. In addition, we tested these methods on multiple datasets for lung cancer with diverse imaging parameters (Tables \ref{Tab1},\ref{Tab2},\ref{Tab3}) and Pneumonia (Fig. \ref{fig4}) to ensure the performance of the approach. We categorize the classification tasks into the following divisions:\\
1) We used various frequently used DR models to reduce HD radiomics as baselines such as the standard Isomap, tSNE, PCA, and PHATE models.\\
2) We conducted binary classification to investigate the power of prediction using LD radiomics of multiple machine learning methods, i.e., random forest, logistic regression, Naïve Bayesian classifier, and kernel support vector machine (SVM). We performed 10-fold cross-validation and fixed hyperparameters to ensure the quality of comparison.\\
3) The Cox proportional hazard (CPH) model is used to predict OS. Kaplan Meier survival curves were also used to distinguish between high- and low- risk patients with respect to median hazard.

\vspace{-.1in}
\subsection{Computational Complexity}
\noindent PR-Isomap changes the complexity of Dijkstra's algorithm on the new graph (see Algorithm \ref{algo:PR-Isomap}), $G_e$, is $O(log(N_e)E_e)$, $N_e$ and $E_e$ are the number of nodes and edges in $G_e$, respectively. For a $q$-nearest-neighbor graph with the parameter $q$, we have $qN/2$ edges, resulting in $N_e = qN/2$ nodes in $G_e$. Connecting each node to all $q$ neighbors, edges in $G_e$ in the worst case scenario is $E_e = ((qN/2)q)/2$, leading to a complexity of $O(log(qN/2)((qN/2)q)/2) = O((q^2N/4)log(qN/2))$. 

\section{Results}
\noindent The proposed method modifies Isomap and examines it by HD throughputs from various datasets. 
\vspace{-.1in}
\subsection{Study population}
\noindent {Diverse datasets comprising CT, PET, and ChestXRay images derived from patients diagnosed with Non-Small Cell Lung Cancer (NSCLC) and Pneumonia were examined to evaluate the proposed approach. There were 70,000 digit image data from the MNIST dataset~\cite{mnist} also used for visualizing the separability of the LD features.}

\textbf{NSCLC Radiogenomics (Multimodal PET/CT):} The first, the NSCLC-Radiogenomics dataset ~\cite{Re29,r30,r31}, is a publicly available dataset from the National Institutes of Health (NIH) — The Cancer Imaging Archive (TCIA), providing images of 211 patients; we analyzed a sub-cohort of 130 patients for whom information on miRNA was available. These provided a subset data of 57836 CT images which had corresponding PET images. Overall, the dataset contains CT and PET images. We used our annotations of the tumors as observed on the medical images using maps of tumors in the CT scans and from the PET scans. This is combined with other clinical information and survival outcomes to predict the survival of the patients. A board-certified thoracic radiologist (M.H.) manually segmented the tumor area using the ITK-SNAP software (version 3.6.0).

The NSCLC Radiogenomics dataset included 20, 79, 27, and 7 patients with slice thickness (ST) of $ST \leq 1, 1 < ST \leq 2, 2< ST \leq 4,$ and $ST \leq 5mm$, respectively; 16 sets of images were acquired with Siemens Healthineers (Erlangen, Germany) scanners, 110 with General Electric Healthcare (Chicago, IL) scanners, 3 with Philips Medical Systems Technologies Ltd. (Amsterdam, Netherlands), and 1 with Toshiba Corporation (Minato, Tokyo, Japan). This dataset had 39 contrast-enhanced and 54 non-contrast-enhanced images (Table \ref{Tab1}).

\begin{figure}[t]
\begin{center}
   \includegraphics[width=0.6\linewidth]{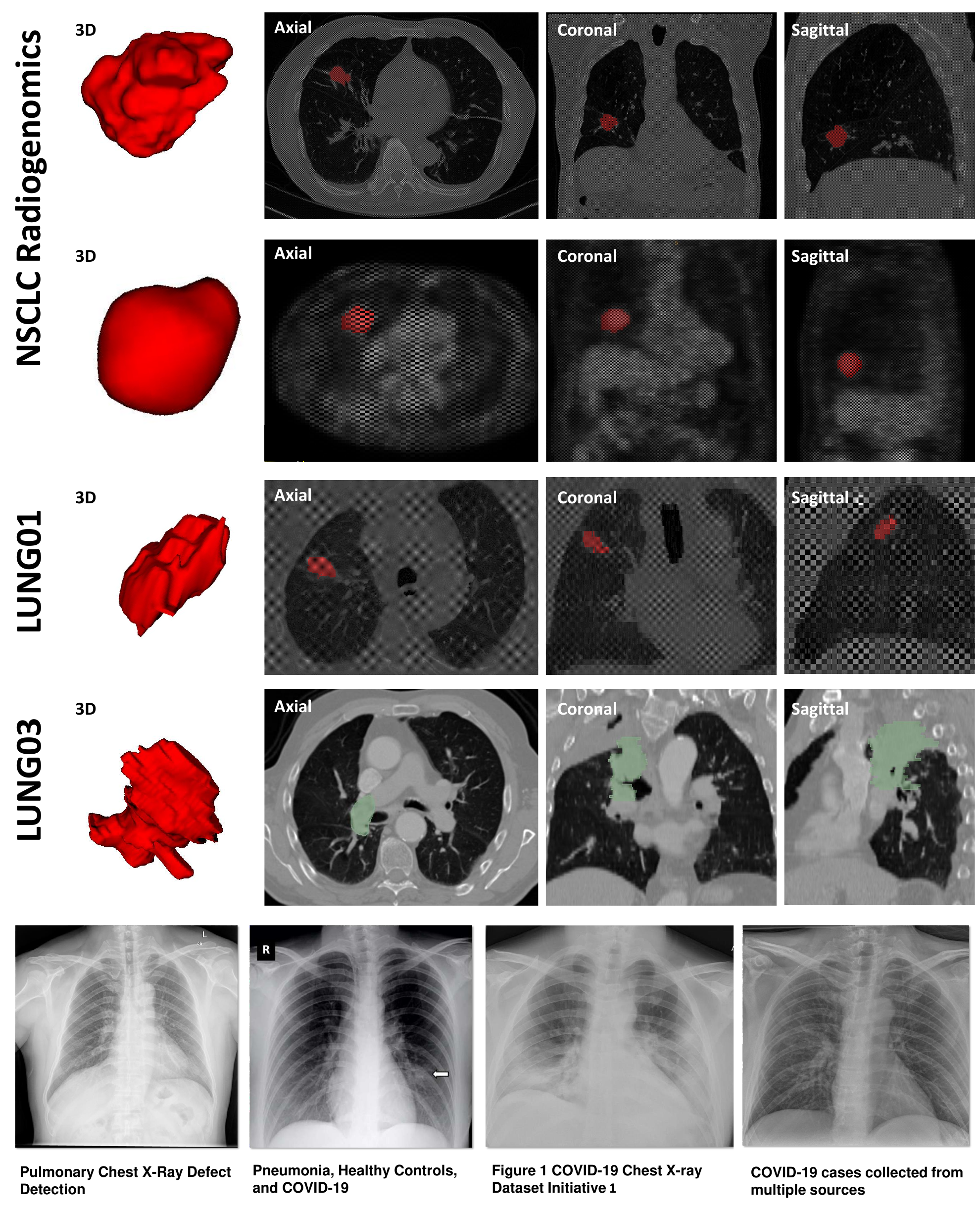}
\end{center}
   \caption{\textbf{Tumors cluster in the two phenotypes.} Visualizations of the original CT/PET images with tumors in the field of view for each dataset with ChestXRay pneumonia data. Tumors’ morphology and shape along with other features affect the quantitative attributes obtained from the selected contours. }
\label{fig4}
\end{figure}

{ \textbf{NSCLC-Radiomics (LUNG01):} The second dataset contains images from 422 non-small cell lung cancer (NSCLC) patients treated at MAASTRO Clinic, The Netherlands. The patients’ pretreatment was conducted using CT scans, and manual delineation by a radiation oncologist of the 3D volume of the total tumor volume with additive clinical outcome data. This dataset refers to the Lung1 dataset and is used to extract high dimensional radiomics and their effects on prediction survival ~\cite{Re32,Re33}.
The LUNG1 dataset comprised 422 patients with $2< ST \leq 4$; 324 sets of images were acquired with Siemens Healthineers (Erlangen, Germany) scanners, and 98 with CMS Inc. Healthcare Solutions (Charleston, SC). This dataset had 157 contrast-enhanced and 264 non-contrast-enhanced images (Tables \ref{Tab2}.}

{ \textbf{NSCLC-Radiomics (LUNG03):} The third dataset contains images and mRNA sequence data used to investigate the association of radiomic imaging features with gene-expression profiles consisting of 88 NSCLC CT scans ~\cite{r33}. The dataset consisted of 60 ($68.1\%$) males and 28 ($31.8\%$) females; 43 ($48.9\%$) patients with adenocarcinoma, 33 ($37.5\%$) with squamous cell carcinoma, and 12($13.4\%$) with another type of NSCLC; 39 ($44.3\%$) patients had stage I disease, 27 ($30.7\%$) had stage II disease, 12 ($13.6\%$) had stage III, and 10 ($11.4\%$) patients had an unknown stage. There were some variabilities in the outcome of the approach while stratifying patients based on CT parameters.
The LUNG3 dataset included 3, 27, and 58 patients with ST of $ST \leq 2, 2 < ST \leq 4$, and $ST \leq 5mm$, respectively; 38 were acquired with Siemens Healthineers (Erlangen, Germany) scanners, 1 with a General Electric Healthcare (Chicago, IL) scanner, and 49 with Philips Medical Systems Technologies Ltd. (Amsterdam, Netherlands). This dataset had 38 contrast-enhanced and 4 non-contrast-enhanced images, where there was no information in the DICOM header files for the rest of these images regarding contrast information (Tables \ref{Tab3}).}

{\textbf{Pneumonia dataset:} A combination of 1125 chest X-ray cases collected, which contains 500 pneumonia cases, 500 no-findings, 125 cases of COVID-19 (Fig. \ref{fig4}). COVID-19 cases were selected from ~\cite{r34} and 1000 images selected from the ChestX-ray8 database ~\cite{r35} contained 500 normal and 500 pneumonia cases ~\cite{r36}. 
55 COVID-19 cases frontal chest X-ray images were selected from Figure 1, DarwinAI and the University of Waterloo have launched an open-source project as a part of the COVIDx dataset to develop our models for COVID-19 detection (COVID-Net) and COVID-19 risk stratification (COVID-RiskNet) ~\cite{r37,r35,r39,r40}. 
All the aforementioned data were collected from multiple sources, which were publicly available databases and online websites such as Figure1.com~\cite{r40}, Radiopaedia.org, the Italian Society of Medical and Interventional Radiology, and the Hannover Medical School, even the images from online publications, websites, or directly from the PDF using the tool pdfimages ~\cite{r41}. All the images were used for different parts of the proposed approach with the overall size of frontal X-rays of $ 512 \times 512$ pixels~\cite{r42}.}

\textbf{MNIST: }The Modified National Institute of Standards and Technology (MNIST) dataset \cite{mnist} is used for the non-biological visualization benchmark of the proposed DR approach. It gives an intuitive understanding of the model’s functionality. MNIST contains 60,000 small square 28×28-pixel grayscale images of handwritten single digits in different scripts between 0 and 9. The pursued task is to inclusively categorize a given handwritten digit while using different DR models into 10 classes indicating 0 to 9 integer values. 



\begin{table*}[]
\centering
\caption[]{The CT parameters for NSCLC Radiogenomics dataset. }
\begin{tabular}{lllll}
\hline
\multicolumn{5}{l}{CT imaging parameters - NSCLC Radiogenomics dataset}                 \\ \hline
Characteristic              & SIEMENS H     & GE HC         & PHILIPS MST   & TOSHIBA C \\ \hline
Contrast Enhanced (CE)           & 5             & 34            & -             & -         \\
Non- CE      & 4             & 50            & -             & 1         \\
Unknown                     & 7             & 26            & 3             &           \\
                            &               &               &               &           \\
Imaging                     & 16            & 110           & 3             & 1         \\
                            &               &               &               &           \\
Convolution Kernel          &               &               &               &           \\
Standard (soft-tissue)      & 2             & 36            & 1 (B,YB)      & -         \\
Hard {\footnotesize(lung, bone, boneplus)} & 14            & 74            & 1 (C)         & 1 (FC52)  \\
                            &               &               &               &           \\
Tube Voltage (kVp)          & 120 (100,140) & 120 (100,140) & 120 (120,140) & 120       \\
Slice Thickness (mm)        &               &               &               &           \\
\textless{}1                & 12            & 7             & 1             & -         \\
1\textless . \textless{}2   & 1             & 77            & -             & 1         \\
2 \textless . \textless{}4  & 3             & 23            & 1             & -         \\
\textless 5                 & -             & 5             & 1             & -         \\ \hline
                            &               &               &               &           \\
                            &               &               &               &          
\end{tabular}
\label{Tab1}
\end{table*}


\begin{table}[tb]
\caption[]{The CT parameters for NSCLC Radiomics-genomics (LUNG01) dataset.}
\centering
\begin{tabular}{lll}
\hline
\multicolumn{3}{l}{CT imaging parameters - NSCLC Radiomics-genomics LUNG01 dataset} \\ \hline
Characteristic                & {\footnotesize SIEMENS H.}   & CMS Inc.              \\
                              &                        & {\footnotesize Healthcare Sol.}  \\ \hline
{\footnotesize Contrast Enhanced (CE)} & 157           & -                     \\
Non- CE                       & 167                    & 97                    \\
Unknown                       & -                      & 1                     \\
                              &                        &                       \\
Imaging                       & 324                    & 98                    \\
                              &                        &                       \\
Convolution Kernel            &                        &                       \\
Standard (soft-tissue)        & 306                    & -                     \\
Hard {\footnotesize(lung, bone, boneplus)}   & 18      & -                     \\
Unknown                       & -                      & 98                    \\
Tube Voltage (kVp)            & 120 (120,140)          & 119(40,360)           \\
Slice Thickness (mm)          &                        &                       \\
$\leq$ 1                        & -                      & -                     \\
1\textless .  $\leq$  2         & -                      & -                     \\
2\textless .  $\leq$ 4          & 324                    & 98                    \\
$\leq$ 5                         & -                      & -                     \\ \hline
\end{tabular}
\label{Tab2}
\end{table}

\begin{table}[]
\centering
\caption[]{The CT parameters for LUNG03 dataset.}
\begin{tabular}{llll}
\hline
\multicolumn{4}{l}{CT imaging parameters - NSCLC Radiogenomics LUNG03 dataset}                     \\ \hline
Characteristic              & SIEMENS H. & GE HC & PHILIPS MST \\ \hline
Contrast Enhanced (CE)           & -                    & -             & 4                             \\
Non- CE      & 37                   & 1             & -                             \\
Unknown                     & 1                    & -             & 45                            \\
                            &                      &               &                               \\
Imaging                     & 38                   & 1             & 49                            \\
                            &                      &               &                               \\
Convolution Kernel          &                      &               &                               \\
Standard (soft-tissue)      & 3                    & -             & 41 (A,B)                      \\
Hard {\footnotesize (lung, bone, boneplus)} & 35                   & 1             & 8 (C,D)                       \\
                            &                      &               &                               \\
Tube Voltage (kVp)          & 120 (120,140)        & 120           & 120 (120,140)                 \\
Slice Thickness (mm)        &                      &               &                               \\
$\leq$ 1                & -                    & -             & -                             \\
1\textless . $\leq$ 2    & -                    & -             & 3                             \\
2 \textless . $\leq$ 4   & 7                    & 1             & 19                            \\
$\leq$ 5                 & 31                   & -             & 27                            \\ \hline

\end{tabular}
\label{Tab3}
\end{table}

\begin{figure}[t]
\begin{center}
   \includegraphics[width=0.75\linewidth]{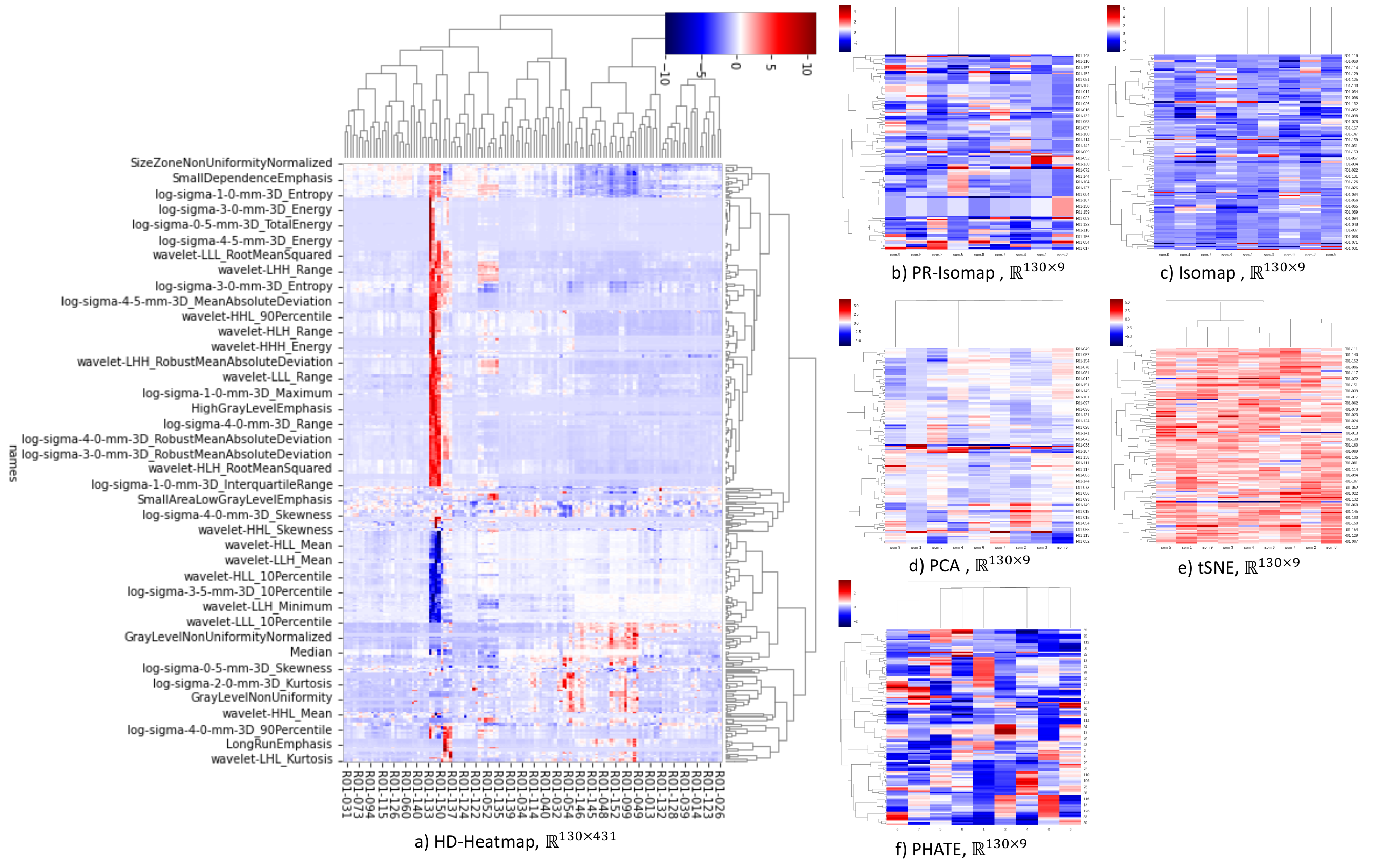}
\end{center}
   \caption{Visualization of the data dimensionality tested on CT radiomics of NSCLC Radiogenomics dataset. a) Representation of HD radiomics, $R^(130×431)$, presented using correlative distance. b) LD demonstration of the HD radiomics projected in the LD space using PR-Isomap. c) Representing the application of HD radiomics in a lower-dimensional space using the original Isomap method.  d) Using PCA, projecting HD features onto LD space. e) t-SNE heatmap to make the same LD imaging biomarkers. }
\label{fig5}
\end{figure}

\vspace{-.1in}
\subsection{Implementation Details}
\noindent The proposed method modifies Isomap and is examined by HD throughputs from handwritten digits and lung cancer datasets. \\
\textbf{Baselines}. We compare PR-constrained Isomap with four baselines,

\begin{itemize}
\item{The standard t-SNE ~\cite{r11}: This method is the standard t-SNE that uses Kullback–Leibler divergence similarity and measures the statistical difference of probability distribution of points in the HD space corresponding to the probability distribution of the same points on the LD space after projection.}

\item{PCA ~\cite{r1}: PCA is a popularly employed method to reduce the dimensionality of data. It captures the maximum variance direction in the data, which maintains the interpretability and maximum amount of information. PCA is a linear method to reduce dimensionality.}

\item{The standard Isomap ~\cite{r3}: Isomap is applied for computing a quasi-isometric and is considered to be a non-linear dimensionality reduction approach, which attempts to maintain the geodesic distances from HD space into the LD space embedding. The standard Isomap does not apply any constraints to maintain the distance between the points.}

\item{PHATE ~\cite{phate}: PHATE emerges as a recently developed DR tool tailored for the visualization of HD data. This method leverages a conceptual framework using the local affinity matrix, Markov transition matrix, and Von Neumann Entropy prior to MDS to acquire and represent the manifold, preserving not only local but also global inter-data point distances. Operating as a visualization technique, PHATE adeptly encapsulates the intricate local and global nonlinear structures within the dataset, achieving this through the application of an information-geometric distance metric interlinking data points.}

\end{itemize}
To visualize the LD embedded data in 3D, we use UMAP ~\cite{r20} to preserve the data’s global structure. 

\begin{figure*} 
    \centering
    \footnotesize
    \setlength{\tabcolsep}{1pt}
    \begin{tabular}{ccccc} 
    \includegraphics[width=0.2\linewidth, angle=0]{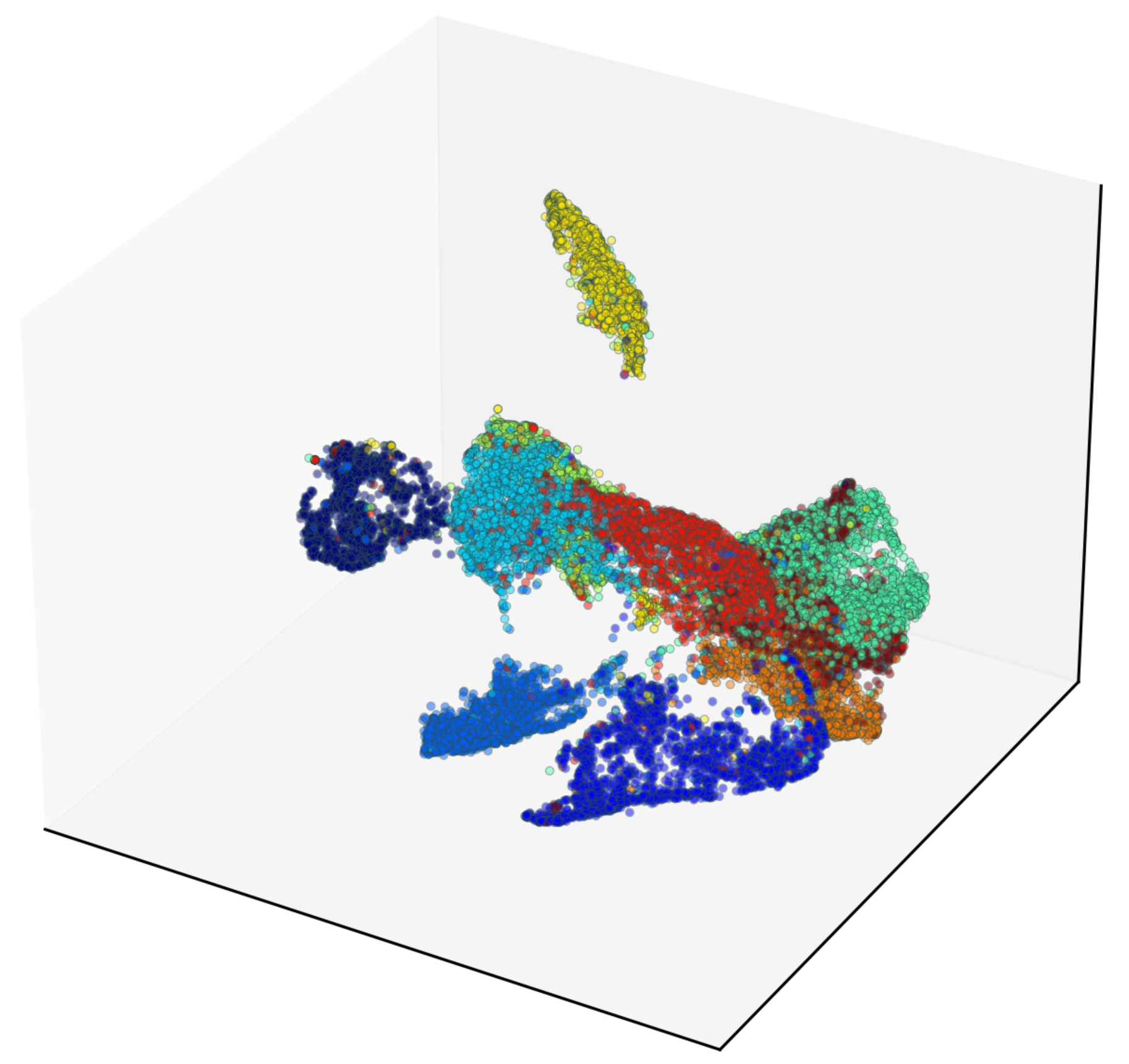} \hspace{-.1cm} & 
    \includegraphics[width=0.2\linewidth, angle=0]{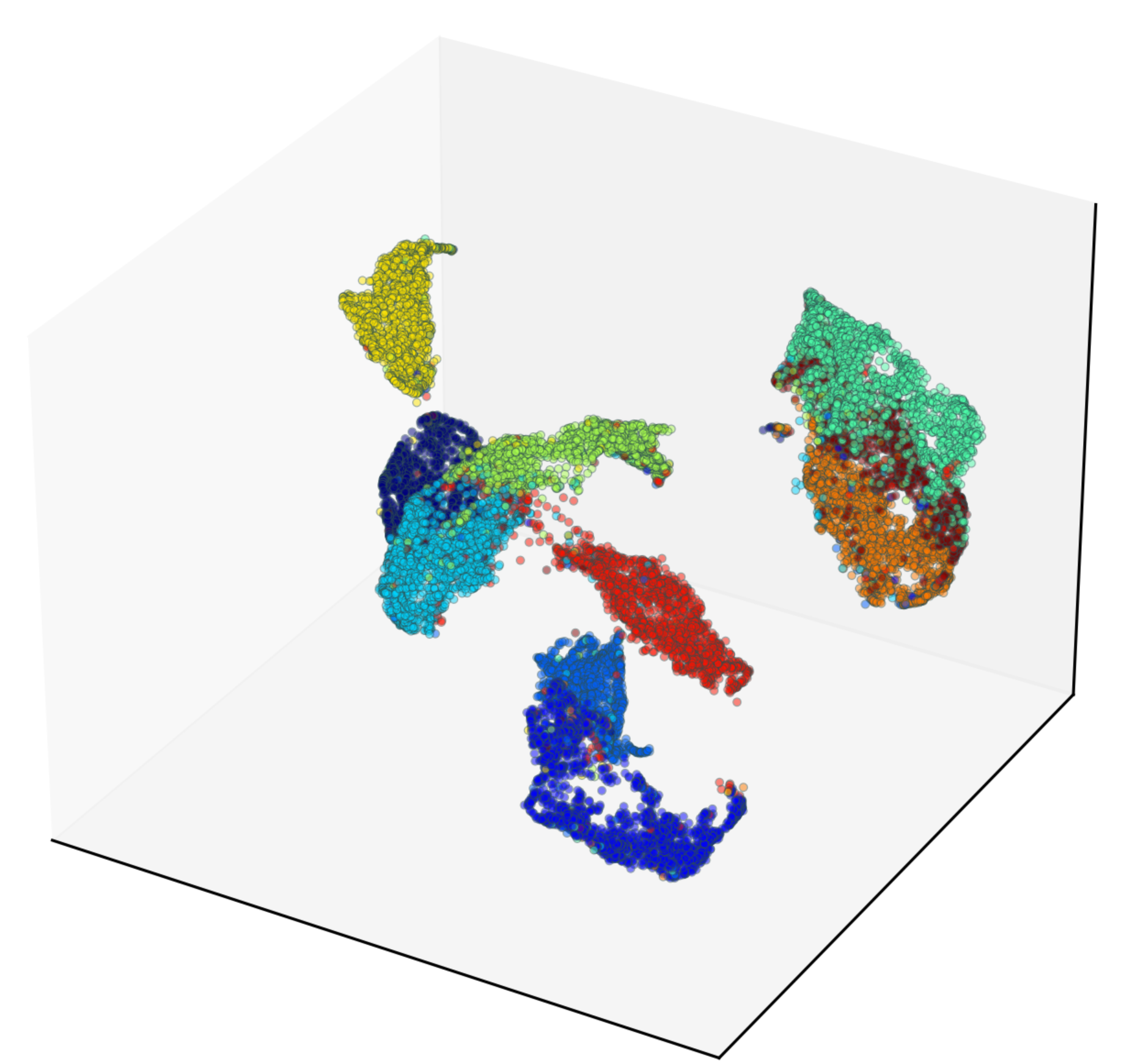}  \hspace{-.1cm} & 
    \includegraphics[width=0.2\linewidth, angle=0]{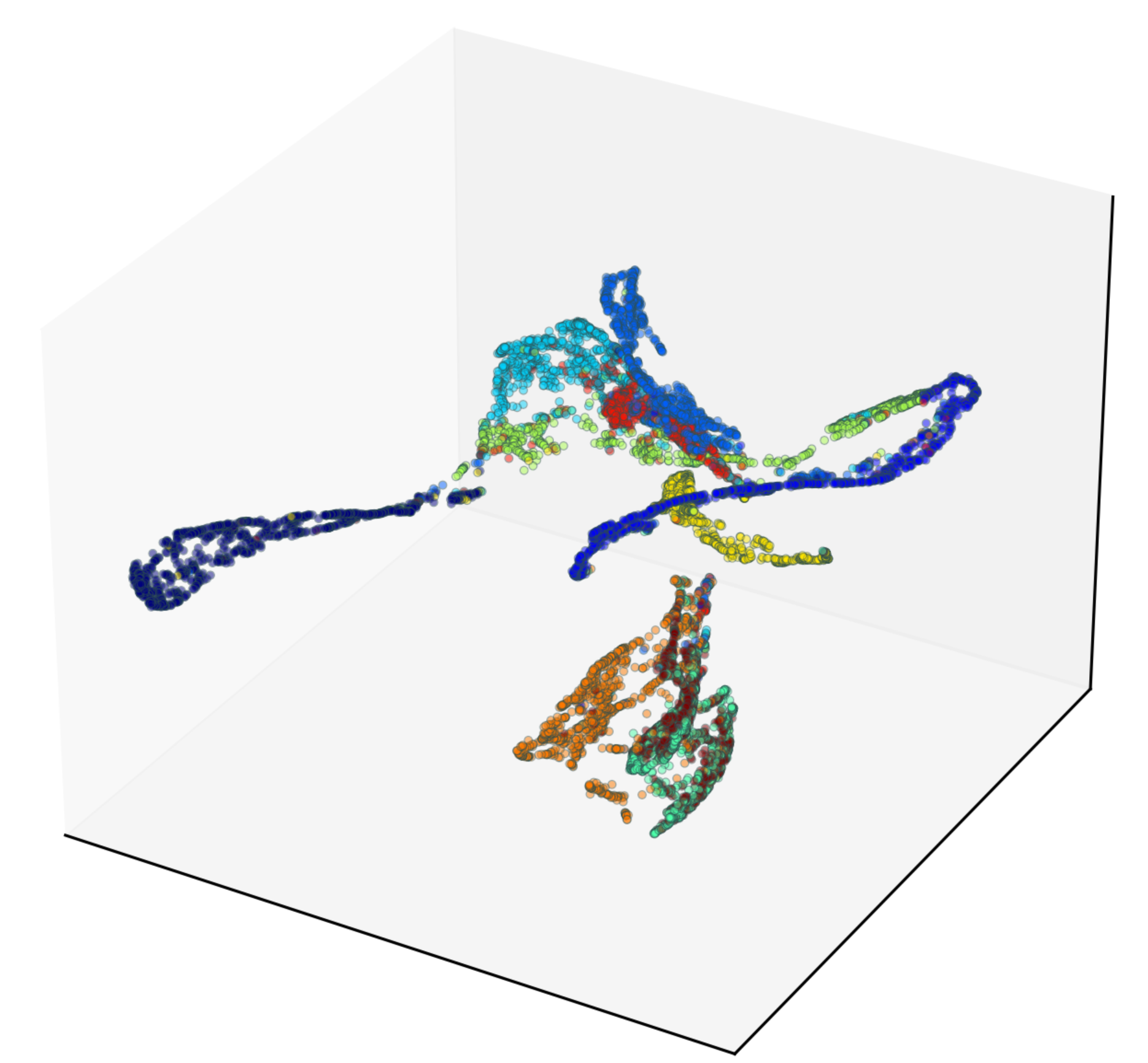}  &     
    \includegraphics[width=0.2\linewidth, angle=0]{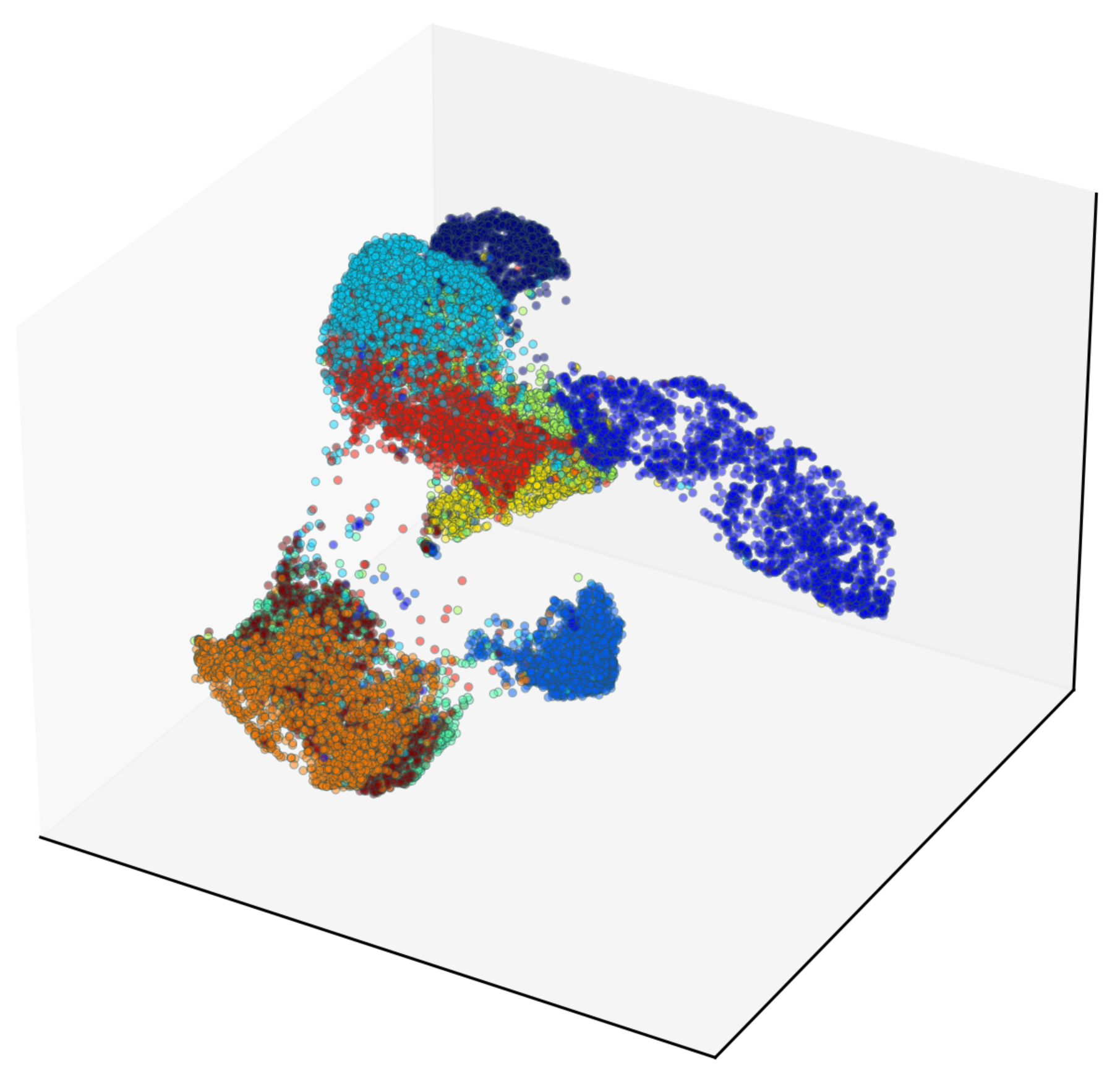} & 
    \includegraphics[width=0.2\linewidth, angle=0]{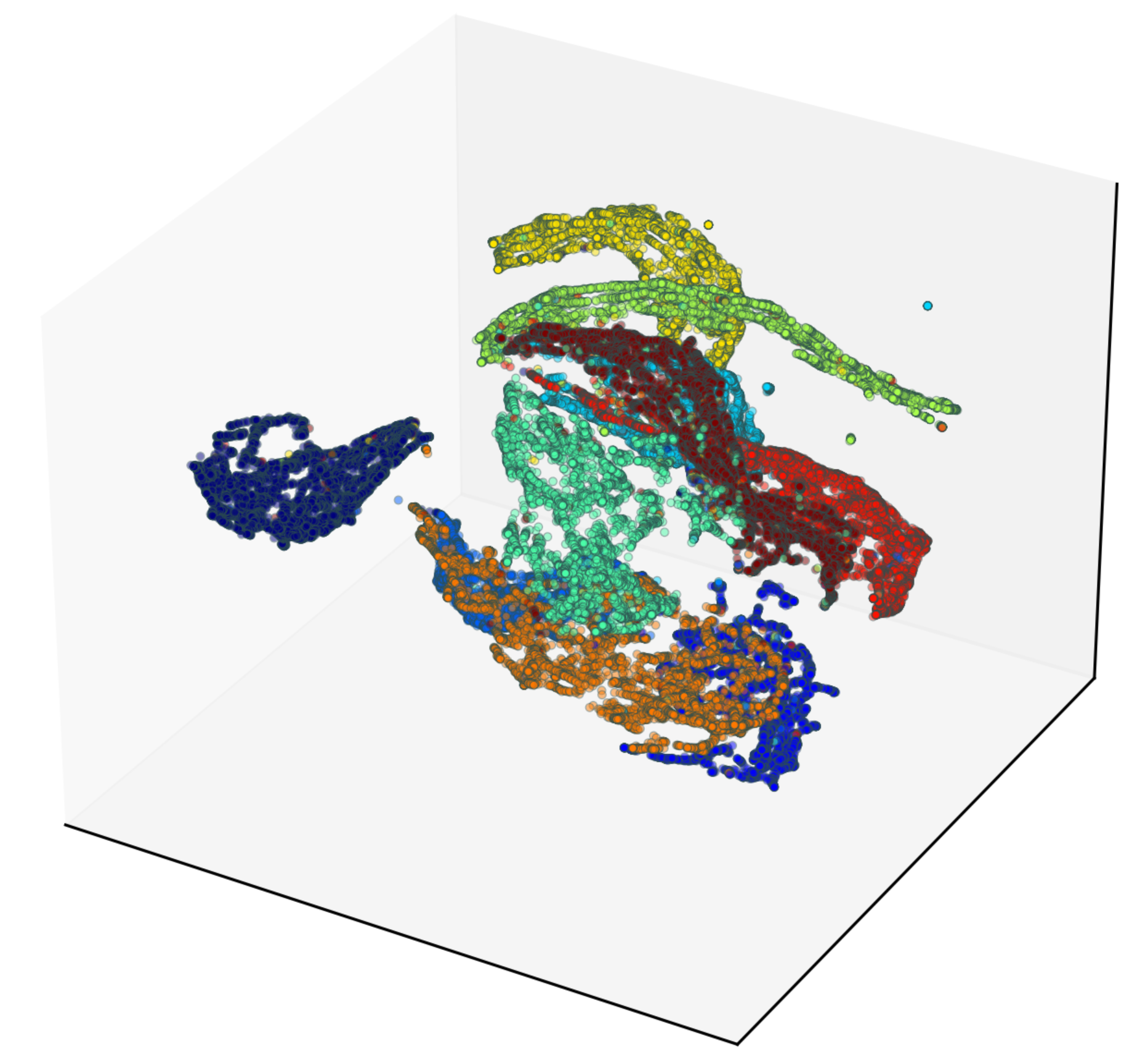}  \\  
       \hspace{-1em}%
    (a) PR-Isomap
    & (b) Isomap
    & (c) tSNE
    & (d) PCA
    & (e) PHATE\\
    \end{tabular}
    \caption{Visualizing the MNIST HD data projected to LD space in 3D space using various DR methods. Mapping with PR-Isomap (\textbf{a}) showed the separation of different categories being closer despite having lesser between-classes (extrinsic or global) distances due to the additive constraint to maintain the uniformity of the classes while intrinsic distances are represented well. The results of PHATE exhibit similar behavior (\textbf{e}) but with lesser strength in grouping the within-class (intrinsic) categories. Isomap (\textbf{b}) and tSNE (\textbf{c}) showed lower uniformity and within-class grouping behavior, respectively.}
    \label{fig:6}
\end{figure*}


\begin{table*}[]
\centering
\caption[]{Predictive power via random forest for five different datasets, overall of 2,236 cases across two different diseases.  }
\resizebox{\columnwidth}{!}{%
\begin{tabular}{llllllll}
\hline
                                     &                      &          &              &             &              &             &             \\ 
Datasets                             & Number of cases      & Modality & PR-Isomap    & Isomap      & PCA          & tSNE        & PHATE       \\
                                     &                      &          & (Ours)       &             &              &             &             \\ \hline
                             &                      &          &              &             &              &             &             \\ 
                                      & {130} & CT       & \textbf{78.5 (±4.4)}  & 76.9(±5.8)  & 73.8 (±7.6)  & 76.9(±4.1)  & 76.2(±7.3)  \\
 {\textbf{NSCLC Radiogenomics} }                                  &                      &          &              &             &              &             &             \\
                                     &                      & PET      & \textbf{78.5 (±5.1)}  & 78.5(±6.3)  & 73.2 (±5.6)  & 76.9(±5.3)  & 75.3(±9.9)  \\
                                     &                      &          &              &             &              &             &             \\
\textbf{LUNG03}               & 88                   & CT       & \textbf{61.4 (±11.4)} & 52.7(±14.8) & 49.6 (±15.1) & 40.3(±15.5) & 45.5(±14.4) \\
                                     &                      &          &              &             &              &             &             \\
\textbf{LUNG01}              & 422                  & CT       & \textbf{88.4 (±1.4)}  & 87.7(±2.8)  & 86.9 (±2.5)  & 87.7(±3.1)  & 88.3 (±0.6) \\
                                     &                      &          &              &             &              &             &             \\
\textbf{Pneumonia}          & 1596                 & Chest X-Ray   & \textbf{80.9 (±5.8)}  & 80.8(±4.8)  & 80.3 (±5.8)  & 79.5(±7.9)  & 78.2(±6.6)  \\
 \textbf{Multiple datasets}                                    &                      &          &              &             &              &             &             \\ \hline

\end{tabular}
}
\label{tab-acc}
\end{table*}

\noindent \textbf{Tasks.} We consider multiple datasets, overall of 72,236 cases, to confirm the efficacy of PR-Isomap: MNIST dataset of handwritten digits, Pneumonia, and NSCLC-Radiogenomics. \\
\textbf{Implementation Details. }To apply PCA, t-SNE, and PHATE, we use the implementation from ~\cite{Re17}, and ~\cite{phate}, respectively. To apply UMAP for LD data visualization, we apply the implementation from ~\cite{r20}. For standard Isomap, we use ~\cite{r3} implemented by Jake Vanderplas (2011). We implement PR-Isomap based on the standard implementation of Isomap by restricting k-nearest neighbors and the shortest path on the manifold and updating it with the PR constraint.\\
\textbf{Hyperparameters. }For PCA and UMAP, we used the default hyperparameters, while we changed the tSNE iteration number to 500 and a random state of 3. For the binary classifiers, we used untuned classification models but we froze the hyperparameters throughout the comparison for different DR methods and all models to keep the integrity of the analysis. In Random Forest, we used 20 estimators, a maximum depth of the forest with four, and a random state of 45. The training follows the standard cross-validation score with k = 8. For PR-Isomap, $\Phi$ was a window function. h\ in the PR function was 16, and 10 for CT and PET radiomics, while it was 70000 in the MNIST dataset and selected based on the lower-bound amount for converting LD features. We used Gap statistics and Elbow to identify the optimum number of LD spaces.
\vspace{-.1in}
\subsection{LD Representation of HD Manifolds}
\noindent Classification evaluation and outcome prediction for the patients screening or under treatment is a popular way of showing the predictability power of LD radiomics. A supervised algorithm, random forest classifier, was used to perform binary prediction of the final survival of the patients to assess the prediction power of LD radiomics generated by the baselines. The model was untuned but constantly across the comparison to perverse the integrity of the evaluations. 
Table \ref{tab-acc} reports the quantitative accuracy produced by PR-Isomap, Isomap, tSNE, PCA, and PHATE. PR-isomap produces better accuracy than other DR techniques. Thus, PR-Isomap can be used to reduce HD radiomics and facilitate outcome predictions.
In the context of visualizing 70,000 observations of MNIST HD manifold projected into the LD 3D space utilizing various DR baseline techniques. Notably, when employing PR-Isomap (\ref{fig:6}-\textbf{a}), it becomes evident that the different categories exhibit closer proximity, despite possessing relatively smaller between-class (global or extrinsic) distances. This can be attributed to the additional PR constraint applied to sustain the uniformity of the class distributions.\\
Moving to the results obtained through PHATE (\ref{fig:6}-\textbf{e}), a somewhat analogous pattern emerges, albeit with a slightly less pronounced ability to effectively group the within-class categories (intrinsic or local). In contrast, when considering Isomap (\ref{fig:6}-\textbf{b}) and t-SNE (\ref{fig:6}-\textbf{c}), we observe distinct characteristics. Isomap appears to exhibit lower uniformity in the distribution of data points, while t-SNE showcases a propensity for within-class categories to cluster more distinctly.\\
These observations collectively shed light on the diverse behaviors exhibited by various DR methods when projecting MNIST HD data into an LD 3D space, emphasizing the utility of PR-Isomap in fostering closer inter-category proximity while upholding the uniformity of class distributions.
\vspace{-.1in}
\subsection{Cox Proportional Hazard Model to Predict Survival}
\begin{figure*}[t]
\begin{center}
   \includegraphics[width=0.9\linewidth]{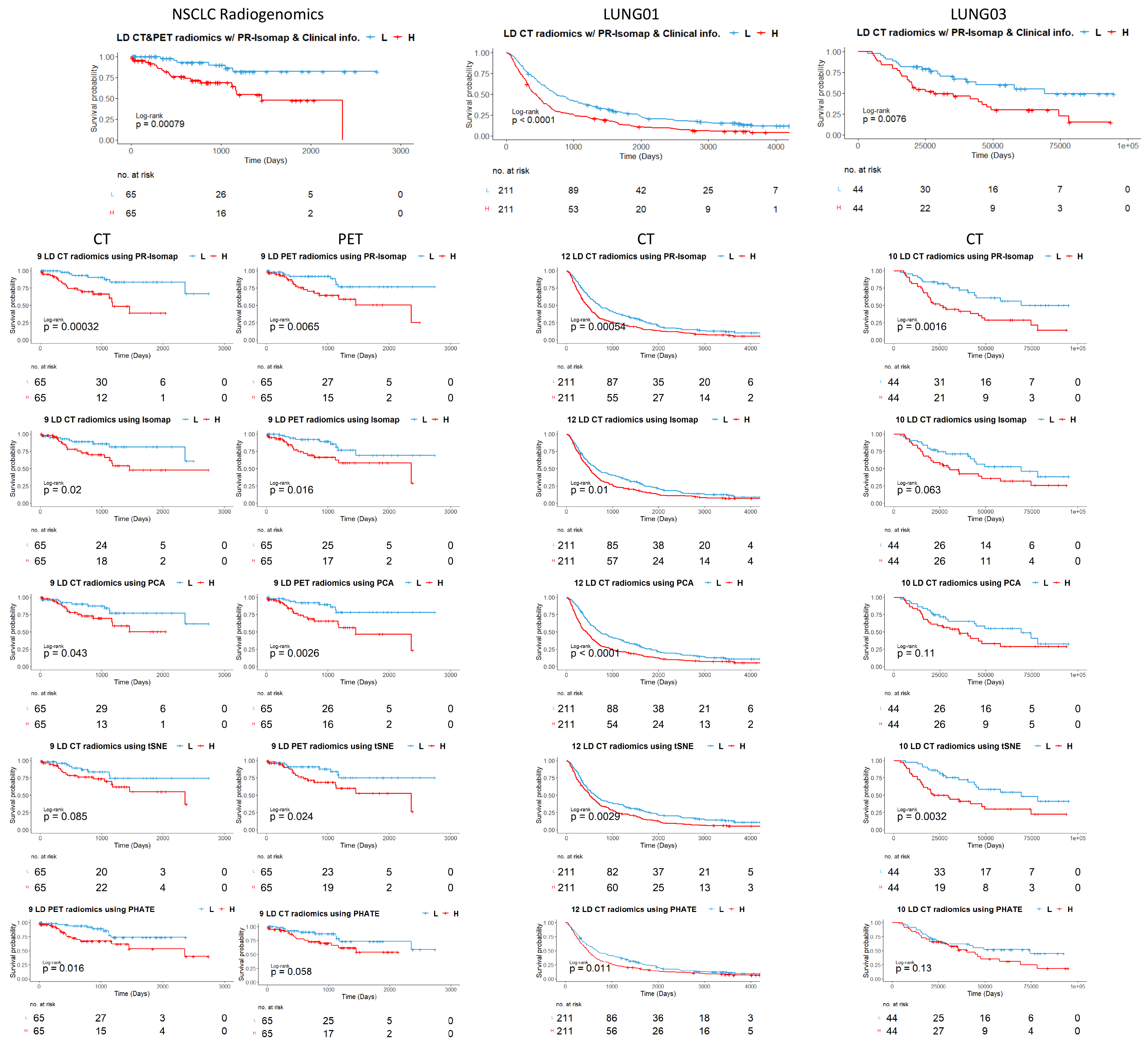}
\end{center}
   \caption{\textbf{LD Multivariant Survival prediction.} The proposed constrained isometrically mapping DR method exhibits impressive discriminatory power in distinguishing between high-risk and low-risk patients based on the median of diverse imaging biomarkers. The incorporation of clinical information and demographic variables using LD radiomics significantly improves the separability of patient groups with distinct risk profiles. This capability holds tremendous potential in accurately predicting patient survival across the course of treatment. }

\label{fig7}
\end{figure*}

\noindent In Cox modeling of OS on the multimodal PET/CT dataset, the C-statistic of CT and PET LD radiomics were 0.68 and 0.67 ($95\%$ CI), with the long-rank likelihood test p-value of $<0.005$, and $0.007$, respectively. This result generated by PR-Isomap DR reduction outperformed other DR methods used for the CT biomarker analysis while maintaining the separation of high- and low- risk patients (Fig. \ref{fig7}). C-statistics of original Isomap while having LD CT and PET radiomics yield 0.66 (p-value = $0.02$) and 0.67 (p-value = $0.02$), then the results of the CT biomarkers considered to be the second-best accuracy after our proposed method. Despite slightly higher C-statistics of PHATE for the NSCLC Radiogenomics dataset, $0.70$ ($0.06$) for CT and $0.72$ ($0.02$) for PET, the separation of high- and low- risk patients under treatment was noticeably lower than other approaches (Fig. \ref{fig7}). 
In PET imaging markers, PCA generates the highest C statistic, which is inconsistent with binary survival prediction. the lowest C statistic belonged to CT-PCA and PET-tSNE radiomics yielding $0.63$ ($0.04$) and $0.6$ ($0.02$), respectively. A model including the CT-tSNE radiomic avatars resulted in C-statistics of $0.64$ with a p-value of $0.09$. For both Lung01 and Lung03 datasets, PR-Isomap generated LD radiomics alone yielded the highest C-Statistics of $0.59$ ($<0.0005$) and $0.66$ ($0.002$), respectively. Also, the Kaplan-Meier survival curve indicates higher separation of the high- and low-risk patients based on the median of the hazard (Fig. \ref{fig7}).
The Cox models of OS all datasets using PR-Isomap radiomic avatars had statistically significant separation of the Kaplan Meier curves for patients above versus below the median hazard compared to other DR methods.

\vspace{-.1in}
\section{Discussion}
\noindent The HD to LD manifold learning, supported by extensive studies~\cite{Re29,r30,r31,Re32,Re33}, proves instrumental in precision medicine, diagnosis, and optimizing treatment planning. This approach effectively addresses the challenges of HD multivariate data overload, notably in 'omics disciplines, by employing DR methods. Such breakthroughs enable the prediction of genetic information directly from CT images and the identification of gene mutations. Additionally, certain imaging features correlate with gene overexpression, a crucial therapeutic target. Given the complexity and dimensionality issues inherent in these models, efficient DR techniques are indispensable. These techniques aid in extracting non-invasive imaging biomarkers and diverse genetic information, ultimately enhancing the precision of targeted therapy and delivering clinical benefits ~\cite{added}. Both genetic and radiomic data often grapple with HD parameters, rendering DR methods indispensable for the precise processing of HD medical data.

Several methods have been employed in previous studies, including PCA, MRMR~\cite{Re6}, clustering, MDS~\cite{r2}, Isomap ~\cite{r3}, LLE~\cite{r4}, tSNE~\cite{Re17}, and PHATE \cite{phate}. These methods project HD data to LD space and share some similarities. We used KL-divergence, which is similarly used in MRMR, for comparison with tSNE. We also considered MDS-based projection by comparing it with standard Isomap and more recently developed PHATE ~\cite{phate}. PR-Isomap can be used to reduce the dimensionality of data points from HD space while preserving the local pairwise distances and structure of the HD data. In addition, Isomap and similarly PR-Isomap exhibit clustering behavior. To measure the strength of the proposed approach, PR-Isomap was compared to unsupervised DR methods with fixed hyperparameters acquired in the tuning phase. Furthermore, our unsupervised model was developed and validated using different independent NSCLC datasets, and comparative analyses were conducted to demonstrate its strength with respect to the state-of-the-art unsupervised feature reduction methods, thereby increasing the novelty of our proposed model.

In precision medicine analysis, LD-projected radiomics models, employing various DR methods, especially PR-Isomap, exhibit the capability to predict outcome events. PR-Isomap consistently outperforms other state-of-the-art unsupervised DR methods with the highest prediction accuracy \ref{tab-acc}. Although the computational complexity of our proposed method aligns with that of the original Isomap, the influence of spatial resolution, contrast, and unbalanced data introduces minor variations in binary classification results. These challenges are addressed using the Cox model and the "time to event" variable, revealing PR-Isomap's superior performance in survival prediction and enhancing the accuracy of survival analysis compared to classification. PR-Isomap surpasses all baseline DR models when utilizing multivariate Cox proportional hazards models. However, in the first NSCLC data, PHATE exhibits slightly higher C-statistics in the NSCLC-Radigenomics dataset. Yet, it struggles with distinguishing high- and low-risk patients using LD radiomic signatures (Fig. \ref{fig7}). Furthermore, the Kaplan-Meier survival curve of PR-Isomap across DR models and different NSCLC datasets demonstrates superior separation between high- and low-risk patients (see Fig. \ref{fig7}).

For prognostic analysis, our validation datasets encompass 510 patients, primarily comprising CT imaging data, with only 130 patients possessing PET radiomics data. Expanding the dataset to include genomic analysis is a potential avenue for future research. Notably, the image sets within the dataset exhibit heterogeneity in acquisition characteristics, such as contrast enhancement, device model and manufacturer, slice thickness, and convolutional kernel. Despite this potential source of noise, we successfully model survival using patient imaging data. Furthermore, the absence of associations between these imaging factors and survival strengthens our results' credibility. Nonetheless, to attain more robust prognostic results, larger and more homogeneous datasets would be advantageous. Additionally, incorporating gene expression data into the DR analysis could enhance the reliability of our proposed method when dealing with HD data.

For diagnostic prediction using LD radiomic signatures, PR-Isomap outperformed other state-of-the-art models, despite a considerably large cohort of patients for pneumonia, more analysis, involving other types of diseases, could help better understand and ensure the performance of the PR-Isomap. Such investigations would provide a more holistic understanding of the model's capabilities and robustness in different clinical scenarios.

It is important to note the limitations of our study. PR-Isomap still suffers from the same weaknesses as the original Isomap, even with the homogeneity constraint, despite the alleviation of this problem. This is due to the nature of the data, which often requires different transformations before projecting it to LD space to overcome the nonuniform structure of data points on the manifold. Other methods like UMAP using differential geometry in HD projection may be beneficial in improving the data processing and interpretability of the results. For that, mapping the HD manifolds to other spaces to redefine uniformity could be an interesting direction to explore.

\vspace{-.1in}
\section{Conclusion}
\noindent This study proposed a novel dimensionality reduction and manifold learning method, PR-Isomap, which preserves information while projecting HD data to LD space using a Parzen-Rosenblatt constraint. PR-Isomap was applied to create LD projection of multiple datasets, overall 72,236 cases, and experimental results showed significant superiority over other commonly used state-of-the-art unsupervised DR methods to predict outcomes or diagnose diseases. PR-Isomap method may be valuable for processing HD medical data, particularly radiogenomics studies. This method has potential applications in radiogenomic-driven diagnosis, treatment planning, and outcome prediction in HD multimodal imaging.



\end{document}